%% file: main.tex
\documentclass{article}

\usepackage[final]{neurips_2024}

\usepackage[utf8]{inputenc} 
\usepackage[T1]{fontenc}    
\usepackage{hyperref}       
\usepackage{url}            
\usepackage{booktabs}       
\usepackage{amsfonts}       
\usepackage{nicefrac}       
\usepackage{microtype}      
\usepackage{xcolor}         

\usepackage{multirow}
\usepackage{enumerate}
\usepackage{xspace}
\usepackage{makecell}
\usepackage{graphicx}

\usepackage{amsmath}
\usepackage{subfigure}

\long\def\EDIT#1{{\color{black}{#1}\color{red}}}

\newcommand{\model}[1]{\texttt{#1}\xspace}
\newcommand{\SCUM}{\model{SiCoUM}}
\newcommand{\ETS}{\model{ETS}}
\newcommand{\CES}{\model{CES}}
\newcommand{\ARIMA}{\model{ARIMA}}
\newcommand{\THeta}{\model{Theta}}
\newcommand{\tik}{\textcolor{red}{\textbf{\textsuperscript{*}}}}

\newcommand{\AutoARIMA}{\model{AutoARIMA}}
\newcommand{\Naive}{\model{Naive}}

\newcommand{\StatsForecast}{\model{StatsForecast}}

\newcommand{\LSTM}{\model{LSTM}}
\newcommand{\MQCNN}{\model{MQCNN}}
\newcommand{\SPADE}{\model{SPADE}}
\newcommand{\DeepAR}{\model{DeepAR}}

\newcommand{\TFT}{\model{TFT}}
\newcommand{\PatchTST}{\model{PatchTST}}

\newcommand{\NBEATS}{\model{NBEATS}}

\newcommand{\TimeGPT}{\model{TimeGPT}}               
\newcommand{\TimesFM}{\model{TimesFM}}               
\newcommand{\Chronos}{\model{ChronosB}}              
\newcommand{\ChronosBS}{\model{ChronosB-S}}           %
\newcommand{\TabPFN}{\model{TabPFN}}                 
\newcommand{\ChronosBolt}{\model{ChronosBolt}}       

\newcommand{\CFA}{\model{CFA}}
\newcommand{\LagLlama}{\model{LagLlama}}             
\newcommand{\MOIRAI}{\model{Moirai}}                 

\newcommand{\dataset}[1]{\texttt{#1}\xspace}
\newcommand{\Mone}{\dataset{M1}}
\newcommand{\Mfour}{\dataset{M4}}
\newcommand{\Mfive}{\dataset{M5}}
\newcommand{\Mthree}{\dataset{M3}}
\newcommand{\Tourism}{\dataset{Tourism}}

\newcommand{\GO}{\dataset{Dataset A}}
\newcommand{\FBA}{\dataset{Dataset B}}
\newcommand{\Mendel}{\dataset{Dataset C}}
\newcommand{\Fresh}{\dataset{Dataset D}}
\newcommand{\ZIPtwo}{\dataset{Dataset E}}
\newcommand{\GlobalRetail}{\dataset{Dataset F}}

\newcommand\btheta{\boldsymbol \theta}
\newcommand{\SE}[1]{\scriptsize \scalebox{.7}{({\color{gray}{#1}})} }

\title{A More Realistic Evaluation of Cross-Frequency \\
Transfer Learning and Foundation Forecasting Models
}

\author{%
  Kin G. Olivares \textsuperscript{*},
  Malcolm Wolff \textsuperscript{*},
  Tatiana Konstantinova \thanks{These authors contributed equally.} \\
  Amazon, New York, USA \\
  \texttt{\{kigutie,wolfmalc,tkonst,\}@amazon.com} \\
  \And
  Shankar Ramasubramanian, Boris Oreshkin, Andrew Gordon Wilson\\
  Amazon, New York, USA \\
  \texttt{\{sramasub,wilsmman,andpotap\}@amazon.com} \\
  \And
  Andres Potapczynski, Willa Potosnak, Michael W. Mahoney, Mengfei Cao, Dmitry Efimov \\
  Amazon, New York, USA \\
  \texttt{\{wpotosna,mfcao,oreshkin,defimov\}@amazon.com} \\
}

\begin{document}

\maketitle

\vspace{.5cm}
\begin{abstract}
    Cross-frequency transfer learning (CFTL) has emerged as a popular framework for curating large-scale time series datasets to pre-train foundation forecasting models (FFMs). Although CFTL has shown promise, current benchmarking practices fall short of accurately assessing its performance. This shortcoming stems from many factors: 
    an over-reliance on miniature-scale evaluation datasets; inadequate treatment of sample size when computing summary statistics; reporting of suboptimal statistical models; and failing to account for non-negligible risks of overlap between pre-training and test datasets. To address these limitations, we introduce a unified reimplementation of widely-adopted neural forecasting networks, adapting them for the CFTL setup; we pre-train only on proprietary and synthetic data, being careful to prevent test leakage; and we evaluate on 15 large, diverse public forecast competition datasets. Our empirical analysis reveals that statistical models' accuracy is frequently underreported. 
    Notably, we confirm that statistical models and their ensembles consistently outperform existing FFMs by more than 8.2\% in sCRPS, and by more than 20\% MASE, across datasets. 
    However, we also find that synthetic dataset pre-training does improve the accuracy of a FFM by 7\%~percent.
\end{abstract}

\vspace{-0.5cm}
\input{sections/01_introduction.tex} 
\input{sections/02_literature_and_methodology.tex} 
\input{sections/03_experiments.tex} 
\input{sections/04_findings_discussion.tex}

\bibliographystyle{plain}
\bibliography{citations.bib}
\medskip

\clearpage
\appendix
\input{appendix/a_literature_review.tex} \clearpage
\input{appendix/b_dataset_details.tex} \clearpage
\input{appendix/b_point_forecast.tex} \clearpage
\input{appendix/c_training_methodology_hypars.tex} \clearpage
\input{appendix/d_note_on_statistical_ensemble.tex} \clearpage
\input{appendix/e_ablation_studies.tex}


\end{document}

%% file: sections/01_introduction.tex
\section{Introduction} \label{01_introduction}


Access to billions of temporal observations offers exciting opportunities for training foundation forecasting models (FFMs); and yet significant challenges remain. 
For example, the method known as cross-frequency transfer learning (CFTL) combines series of measurements at different frequencies to train global models~\citep{vanness2023crossfrequencytimeseriesmetaforecasting, salesforce2023moirai}; and, as such, it is an intuitive approach to increase time series dataset sizes. 
As shown in Figure 1, a key challenge in CFTL is the imbalance of observations across series: high-frequency series vastly outnumber lower-frequency ones, causing the model to become saturated and dominated by abundant high-frequency data. Similarly, differences in scale across series bias gradient updates toward larger-scaled series, preventing the model from learning a common representation that performs well across all scales.

Recent work has suggested that zero-shot CFTL can significantly outperform both traditional statistical models as well as full-shot neural forecast models trained on frequency-specific data. 
\TabPFN~\citep{hutter2025tabpfn}, \TimesFM~\citep{google2024timesfm}, \model{Chronos}~\citep{aws2024chronos}, \MOIRAI~\citep{salesforce2023moirai} researchers report improvements of over 35\% in probabilistic forecasting accuracy compared to traditional approaches like \ARIMA~\citep{hyndman2008automatic_arima} and statistical ensembles, and more than 15\% relative to smaller deep learning architectures such as \NBEATS~\citep{oreshkin2020nbeats}. 
However, practical adoption of FFMs as out-of-the-box replacements for statistical or frequency-specific neural forecast models remains low, and forecasting practitioners have questioned the validity of these improvement claims and the experimental conditions under which they were obtained~\citep{nixtla2025chronos}.

\begin{figure}[t] 
    \centering
    \subfigure[Unbalanced Observations]{
    \label{fig:unbalanced_observations}
    \includegraphics[width=0.45\linewidth]{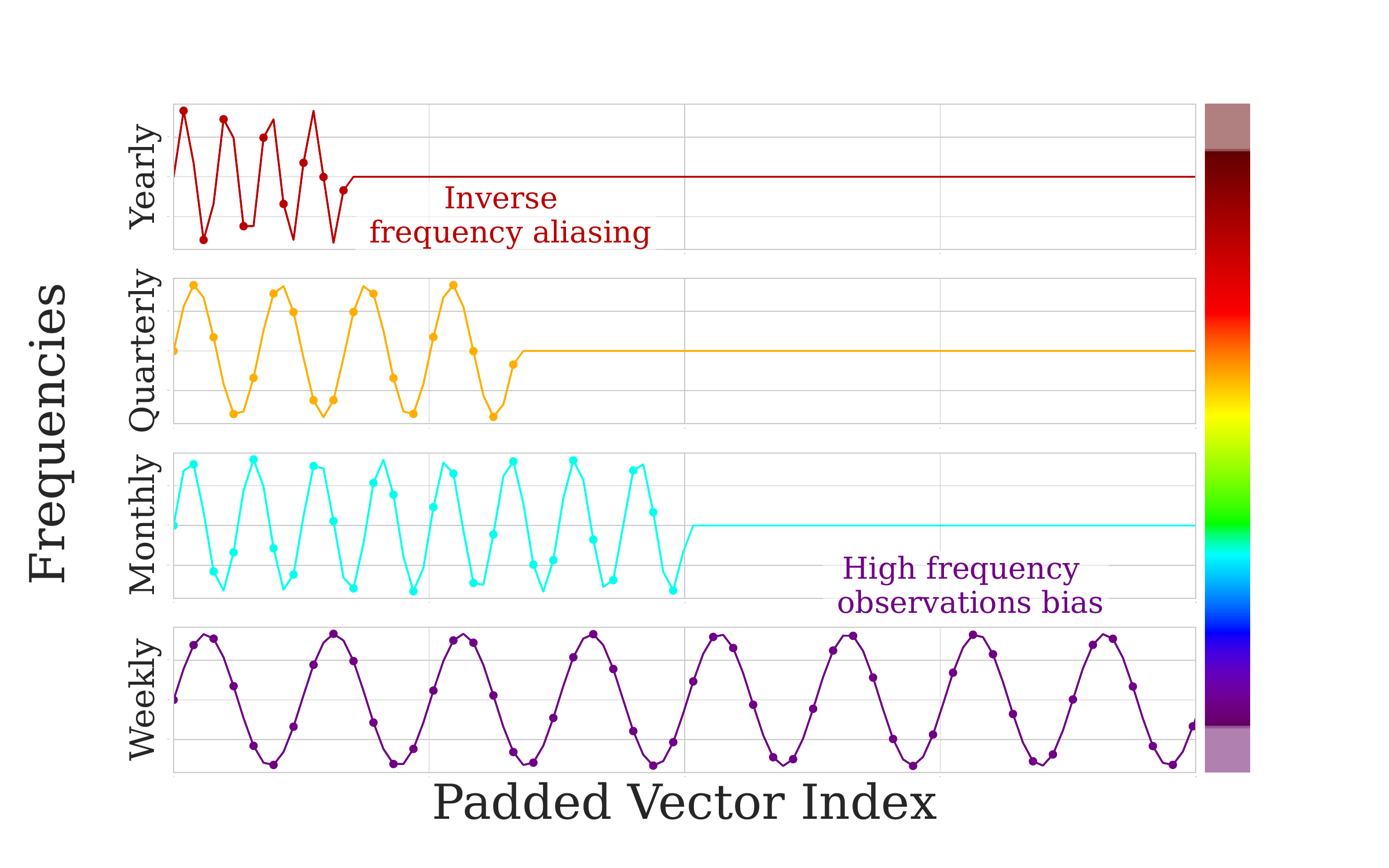}
    }
    \subfigure[Heterogeneous Scales]{
    \label{fig:heterogeneous_scales}
    \includegraphics[width=0.48\linewidth]{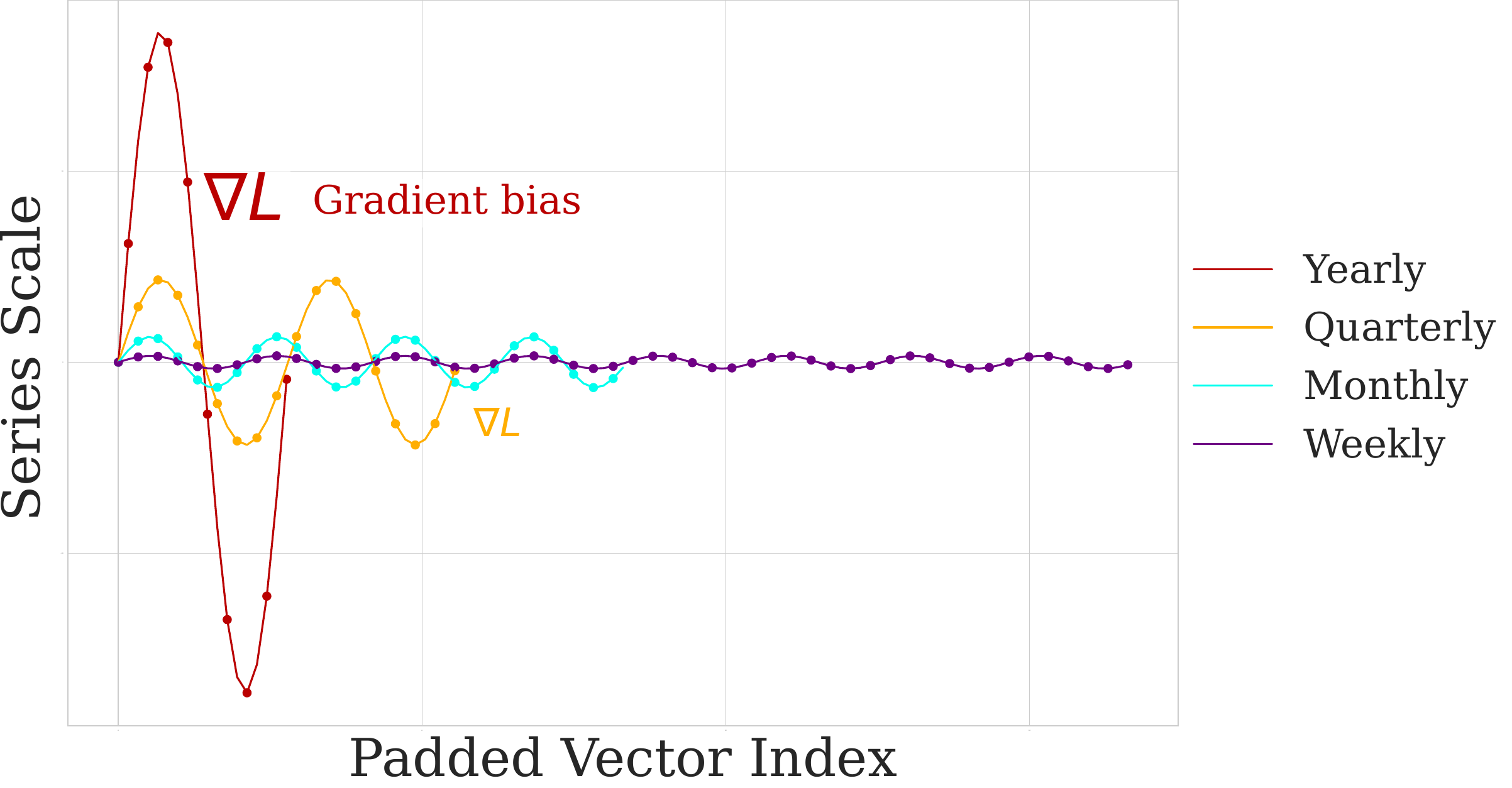}
    }
    \caption{
    \EDIT{Naively padding and combining series of different frequencies to train global models leads to two challenges:
    (a) the unbalanced observations of series of different frequencies, saturate learning signals and induce inverse frequency aliasing effects; and (b) heterogeneous time series scales, that bias gradient optimization.}
    These unresolved challenges still prevent FFMs to replace statistical models and neural forecasting models specialized on each frequency.
    }
    \label{fig:motivation}
\end{figure}

\EDIT{
In this paper we argue that the appropriate criterion for assessing the success of CFTL FFMs is their ability to outperform well-established, frequency-specialized statistical models in zero-shot settings that closely resemble the conditions under which practitioners operate. We also explore the question: \textit{Are current celebrations of CFTL’s superiority over statistical methods premature?}
}


Our key contributions include the following.
\begin{enumerate}[(i)]
    \item 
    \textbf{Unified CFTL Framework.} 
    We re-implement a collection of well-established neural forecasting models and adapt them to share \EDIT{optimization, forecast outputs and evaluation pipeline}. \EDIT{Our framework} enables controlled comparisons by standardizing the pre-training data, model estimation strategy, model outputs, and hyperparameter tuning budget.
    \item 
    \textbf{Careful Pre-Train Dataset Curation.} 
    To prevent any test data leakage in our transfer learning task, we pre-train exclusively on \EDIT{proprietary} and synthetic datasets, and evaluate on 15 large-scale forecasting competition datasets. Our pre-train corpus comprises over 1.58 billion time series, spanning frequencies from daily to yearly. We further demonstrate that, even with extensive proprietary data, the inclusion of simple synthetic datasets improves CFTL's sCRPS accuracy by 7\%. and MASE by 20\%.

    \item 
    \textbf{Fair Comparison of CFTL and statistical models.} We benchmark our FFMs against automatic statistical models~\citep{hyndman2025forecasting}, and \EDIT{ensure their specialization in each series, by properly defining its hyperparameter search space based on their frequency}. Furthermore, rather than relying solely on aggregate metrics - which can bias the evaluation toward smaller datasets - we report disaggregated results and use weighted averages to provide a more balanced and representative assessment across datasets. We release the evaluation of our statistical models at \url{https://anonymous.4open.science/r/neurips_baselines-4BC5}. 
\end{enumerate}

The paper is structured as follows: 
Section~\ref{02_methodology} introduces the CFTL methodology and reviews relevant literature; 
Section~\ref{03_experiments} presents our main experiments and summarizes our main empirical findings;
and 
Section~\ref{04_findings_discussion} concludes. 

%% file: sections/02_literature_and_methodology.tex
\section{Methodology} \label{02_methodology}

We 
consider the univariate forecasting task. 
Let's start by introducing its mathematical notation. Let the forecast creation dates be $[t]=[1,...,T]$ and the forecast horizon be denoted by $[h]=[1,2,...,H]$. 
Given a time series target variable $\mathbf{y}=\mathbf{Y}_{[t][h]}$ and target history $\mathbf{y}_{[:t]}$, the forecasting task estimates the following conditional probability:

\begin{equation}
\mathbb{P}\left(\mathbf{Y}_{[t][h]}\,\mid\,
\btheta,\;
\mathbf{y}_{[:t]}\right).
\label{eq:normal_forecast}
\end{equation}

\paragraph{Model Estimation.} 
Consider a source forecast dataset $D^{(S)}$, defined as the set of realization tuples $D^{(S)}=\{(\mathbf{x},\mathbf{y})|\; \mathbf{x} \in \mathcal{Y}_{[t-L:t]},\; \mathbf{y} \in \mathcal{Y}_{[t][h]}\}$, where the  $\mathcal{Y}_{[t][h]}$ and $\mathcal{Y}_{[t-L:t]}$ are the target variable and regressor support space. We estimate each forecasting model parameters $\btheta$ by minimizing the empirical risk based on Quantile Loss (QL; \citep{koenker_bassett1978quantile_regression}).

\begin{figure}[!t] 
    \centering
    \includegraphics[width=0.85\linewidth]{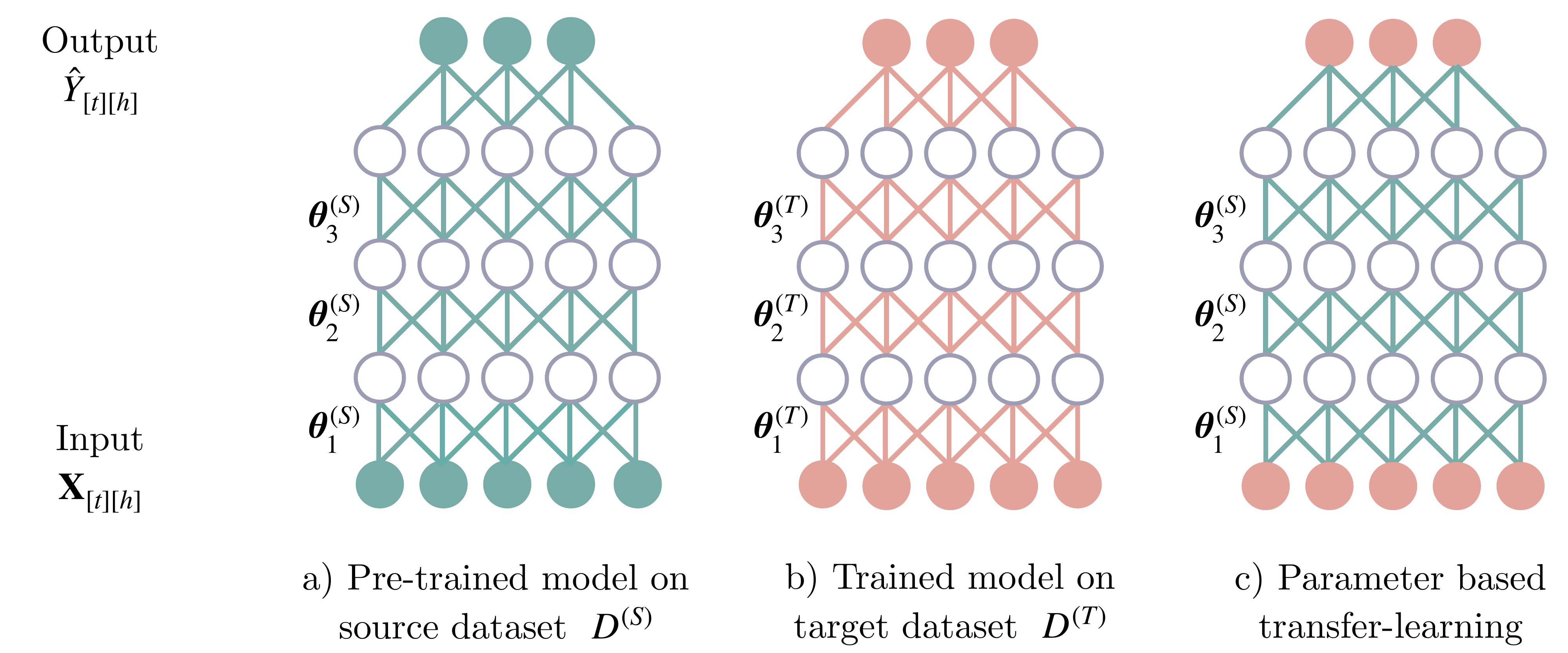}
    \caption{Three-layer fully connected network predictive function. Classic forecasting applications optimize distinct model parameters for source $D^{(S)}$ and target $D^{(T)}$ datasets, a) and b) columns. Parameter-based transfer-learning leverages source dataset knowledge by using a pre-trained model's parameters $\boldsymbol{\theta}^{(S)}_{l}$, to initialize another model's parameters $\boldsymbol{\theta}^{(T)}_{l}$ that can specialize on a target dataset.} \label{fig:transfer_intuition}
\end{figure}

\paragraph{Transfer Learning Forecasting Task.} As shown in Figure~\ref{fig:transfer_intuition}, the zero-shot forecast task distinguishes the source data sets $D^{(S)}$ and target $D^{(T)}$, the task indirectly uses the information from the source domain by using the transfered parameters~\citep{yosinski2014original_transfer} as the forecasting function  from Equation~\eqref{eq:normal_forecast}. Literature review for forecasting transfer learning is available in Appendix~\ref{appendix:literature_review}.

%% file: sections/03_experiments.tex
\section{Experiments} \label{03_experiments}



\paragraph{Pre-train Datasets.} 
To pretrain our models, we use a diverse collection of 1.58 billion large online retail time series spanning daily, weekly, monthly, quarterly, and yearly frequencies. 
These datasets include demand data from cashierless convenience stores, grocery delivery services, and physical grocery stores.
We augment the large-scale online retail demand data with a synthetic dataset composed using a combination of Fourier harmonic signals to mimic seasonalities, polynomial trends, Gaussian processes that we depict in Figure~\ref{fig:training_methodology}. \EDIT{Dataset details in Appendix~\ref{appendix:dataset_details}.}

\paragraph{M-series Evaluation Datasets.}
We consider 15 large scale forecast datasets comprising over 100,000 time series, curated from major forecasting competitions: \Mone~\citep{makridakis1982m1_competition}, \Mthree~\citep{makridakis2000m3_competition}, \Mfour~\citep{makridakis2020m4_competition}, and \Tourism~\citep{athanosopoulus2011tourism_competition}. 
These datasets, represent a broad range of domains and temporal frequencies. 
To ensure comparability with recent neural forecasting literature, we adopt the data handling and pre-processing practices of \Chronos~\citep{alexandrov2020package_gluonts,aws2024chronos} and \NBEATS~\citep{oreshkin2020nbeats}. 
Importantly, we use the datasets solely for evaluation purposes -- excluding them from model optimization -- to assess their true zero-shot forecasting capabilities of our models and avoid any potential test leakage.


\paragraph{Forecasting Baselines.}  
We compare FFMs with two statistical models: \AutoARIMA~\citep{box_jenkins1970book, hyndman2008automatic_arima}, and the Simple Combination of Univariate Models (\SCUM, \citep{petropoulos2020scum}), using the \StatsForecast library~\citep{garza2022statsforecast, hyndman2025forecasting}.
Details of the implementation can be found in Appendix~\ref{appendix:statistical_ensemble}. 
In addition, we consider the following neural forecasting baselines \NBEATS~\citep{oreshkin2020nbeats,olivares2022nbeatsx} , \MQCNN~\citep{wen2017mqrcnn,olivares2021pmm_cnn}, \PatchTST~\citep{yuqi2023patchtst}, \ChronosBolt~\citep{aws2024chronos}, and \MOIRAI~\citep{salesforce2023moirai}. 
Details on the unified CFTL framework implementation and hyperparameters are available in Appendix~\ref{appendix:training_methodology}.



Although we are unable to control hyperparameters or ensure the zero-shot regime (as external FFMs used M-competitions to train), we still evaluate external FFMs from the original \TabPFN~\citep{hutter2025tabpfn}, \MOIRAI~\citep{salesforce2023moirai}, \TimesFM~\citep{google2024timesfm}, and \Chronos~\citep{aws2024chronos} publications. 
Regarding \MOIRAI, reasonable accuracy requires manual selection of patch sizes and context lengths, as the automatic heuristic frequently leads to catastrophic results. 
For longer daily/weekly series, the memory footprint scaled unfavorably with sequence length, causing out-of-memory failures even with batch size 1, which prevented us from evaluating daily and weekly settings with constant contexts.



We evaluate the accuracy of the forecasts using \emph{scaled Continuous Ranked Probability Score} (sCRPS, \citep{gneiting2007strictly}), defined as follows: 
\begin{equation}
    \mathrm{sCRPS}\left(\mathbf{y},\; \hat{\mathbf{Y}} \right) 
    = 
    \frac{\sum_{i,t,h} \mathrm{CRPS}(y_{i,t,h},\hat{Y}_{i,t,h}) }
    {\sum_{i,t,h} | y_{i,t,h} |} .
\label{eq:scrps}
\end{equation}

We use a Riemann integral approximation technique that uniformly averages the quantile loss over a discrete set of quantiles.
\begin{equation}
\begin{aligned}
    \mathrm{CRPS}(y,\hat{Y}) &= 2\int \text{QL}_q(y,F^{-1}_{\hat{Y}}(q))dq, \\
    \mathrm{where} \qquad \text{QL}_q(y,F^{-1}_{\hat{Y}}(q)) &= q(y-F^{-1}_{\hat{Y}}(q))_+ + (1-q)(F^{-1}_{\hat{Y}}(q)-y)_+.
\end{aligned}
\label{eq:crps}
\end{equation}

\input{tables/main_crps.tex}

\clearpage
\paragraph{Summary of Results.}
Table~\ref{table:main_crps} shows that \ARIMA and \SCUM consistently outperform FFMs, achieving the lowest errors across for 11 of 15 datasets. Neural architectures occasionally match or surpass the baselines, but they never achieve the best score across all frequencies of an entire competition. We complement Table~\ref{table:main_crps} with point forecast evaluations using the \emph{mean average scaled error} (MASE), reported in Appendix~\ref{appendix:point_forecast}. 

Overall, confirming observations from the forecasting community~\citep{nixtla2025chronos}, and in contrast to recent claims of major advances over statistical models~\citep{salesforce2023moirai, aws2024chronos, hoo2024tabpfn, google2024timesfm}, our results show that \ARIMA and \SCUM outperform CFTL-FFMs in both probabilistic and point forecasting tasks. 
Excluding \TimesFM (non-zero-shot), the statistical models and best FFMs performance differs by 8.2\% in weighted sCRPS and by 20\% in weighted MASE. 

%% file: tables/main_crps.tex
\begin{table}[!t]
\caption{Empirical evaluation of probabilistic forecasts. Mean \emph{scaled continuous ranked probability score} (sCRPS) averaged over 5 runs. The best united CFTL framework result is highlighted (lower measurements are preferred). 
The methods without standard deviation have deterministic solutions.
\\
\tiny{
\tik \Chronos-S stands for the pretrained \texttt{ChronosBolt-Small}. Zero-shot predictions correspond to the original Hugging face model published by Fatir et al~\citep{aws2024chronos}.
\tik \Chronos is trained in our unified CFTL framework, without being full-shot we are able to replicate or improve \ChronosBS accuracy in various datasets.\\
\tik\tik Neither \TimesFM nor \ChronosBS are zero-shot forecasting models as they are trained on the \Mfour dataset~\citep{aws2024chronos, google2024timesfm}.
}
} \label{table:main_crps}
\centering
\scriptsize
\resizebox{\textwidth}{!}{\begin{tabular}{ll|cc|ccccc|cccc}
\toprule
                                &           & \multicolumn{2}{|c|}{StatsForecast}  & \multicolumn{5}{c}{Unified CFTL framework}            & \multicolumn{4}{|c}{External FFMs (not zero-shot)}                                                           \\ \midrule
                                & Freq      & \ARIMA          & \SCUM             & Best             & \NBEATS         & \MQCNN            & \PatchTST      & \Chronos\tik  & \MOIRAI-S     & \TabPFN       & \ChronosBS\tik   & \TimesFM\tik\tik         \\ \midrule
\multirow{6}{*}{\Mone}          & M         & 0.154           & 0.168             & \textbf{0.152}   & 0.152           & 0.155             & 0.156          & 0.156         &    0.135      & 0.168         & 0.173            & 0.130                    \\
                                &           & \SE{-}          & \SE{-}            & \SE{-}           & \SE{0.014}      & \SE{0.001}        &  \SE{0.003}    & \SE{0.008}    &    \SE{-}     & \SE{0.003}    & \SE{-}           & \SE{-}                   \\
                                & Q         & 0.088           & 0.084             & \textbf{0.083}   & 0.087           & 0.083             & 0.107          & 0.133         &    0.077      & 0.095         & 0.084            & 0.113                    \\
                                &           & \SE{-}          & \SE{-}            & \SE{-}           & \SE{0.015}      & \SE{0.001}        &  \SE{0.007}    & \SE{0.024}    &    \SE{-}     & \SE{0.0114}   & \SE{-}           & \SE{-}                   \\
                                & Y         & 0.133           & \textbf{0.129}    & 0.134            & 0.151           & 0.182             & 0.137          & 0.163         &    0.210      & 0.143         & 0.119            & 0.145                    \\ 
                                &           & \SE{-}          & \SE{-}            & \SE{-}           & \SE{0.016}      & \SE{0.022}        &  \SE{0.011}    & \SE{0.023}    &    \SE{-}     & \SE{0.012}    & \SE{-}           & \SE{-}                   \\ \midrule
\multirow{8}{*}{\Mthree}        & O         & 0.034           & \textbf{0.034}    & 0.045            & 0.052           & 0.045             & 0.073          & 0.077         &    0.035      & 0.038         & 0.036            & 0.040                    \\
                                &           & \SE{-}          & \SE{-}            & \SE{-}           & \SE{0.021}      & \SE{0.008}        &  \SE{0.010}    & \SE{0.03}     &    \SE{-}     & \SE{0.008}    & \SE{-}           & \SE{-}                   \\
                                & M         & 0.098           & \textbf{0.095}    & 0.104            & 0.111           & 0.117             & 0.105          & 0.104         &    0.093      & 0.107         & 0.113            & 0.089                    \\
                                &           & \SE{-}          & \SE{-}            & \SE{-}           & \SE{0.0010}     & \SE{0.008}        &   \SE{0.002}   & \SE{0.004}    &    \SE{-}     & \SE{0.001}    & \SE{-}           & \SE{-}                   \\
                                & Q         & 0.077           & \textbf{0.073}    & 0.080            & 0.083           & 0.080             & 0.103          & 0.121         &    0.077      & 0.077         & 0.074            &  0.075                   \\
                                &           & \SE{-}          & \SE{-}            & \SE{-}           & \SE{0.016}      & \SE{0.009}        &  \SE{0.006}    & \SE{0.025}    &    \SE{-}     & \SE{0.005}    & \SE{-}           & \SE{-}                   \\
                                & Y         & 0.156           & 0.144             & \textbf{0.127}   & 0.127           & 0.167             & 0.129          & 0.156         &    0.135      & 0.132         & 0.114            & 0.144                    \\
                                &           & \SE{-}          & \SE{-}            & \SE{-}           & \SE{0.012}      & \SE{0.017}        &  \SE{0.008}    & \SE{0.020}    &    \SE{-}     & \SE{0.007}    & \SE{-}           & \SE{-}                   \\ \midrule
\multirow{10}{*}{\Mfour}        & D         & 0.024           & 0.024             & \textbf{0.023}   & 0.077           & 0.023             & 0.021          & 0.019         &    0.033      & 0.023         & 0.028            & 0.021                    \\
                                &           & \SE{-}          & \SE{-}            & \SE{-}           & \SE{0.003}      & \SE{0.001}        & \SE{0.001}     & \SE{0.001}    &    \SE{-}     & \SE{0.001}    & \SE{-}           & \SE{-}                   \\
                                & W         & \textbf{0.046}  & 0.049             & 0.047            & 0.067           & 0.047             & 0.050          & 0.050         &    0.071      & 0.046         & 0.053            & 0.042                    \\
                                &           & \SE{-}          & \SE{-}            & \SE{-}           & \SE{0.002}      & \SE{0.005}        &  \SE{0.002}    & \SE{0.001}    &    \SE{-}     & \SE{0.001}    & \SE{-}           & \SE{-}                   \\
                                & M         & 0.096           & \textbf{0.096}    & 0.101            & 0.105           & 0.108             & 0.095          & 0.097         &    0.117      & 0.101         & 0.108            & 0.066                    \\
                                &           & \SE{-}          & \SE{-}            & \SE{-}           & \SE{0.001}      & \SE{0.004}        &  \SE{0.002}    & \SE{0.003}    &    \SE{-}     & \SE{0.001}    & \SE{-}           & \SE{-}                   \\
                                & Q         & 0.079           & \textbf{0.078}    & 0.085            & 0.090           & 0.085             & 0.092          & 0.081         &    0.151      & 0.084         & 0.080            & 0.062                    \\
                                &           & \SE{-}          & \SE{-}            & \SE{-}           & \SE{0.001}      & \SE{0.005}        &  \SE{0.005}    & \SE{0.002}    &    \SE{-}     & \SE{0.002}    & \SE{-}           & \SE{-}                   \\
                                & Y         & 0.125           & \textbf{0.115}    & 0.133            & 0.133           & 0.159             & 0.121          & 0.144         &    0.187      & 0.121         & 0.106            & 0.091                    \\
                                &           & \SE{-}          & \SE{-}            & \SE{-}           & \SE{0.010}      & \SE{0.017}        &  \SE{0.010}    & \SE{0.019}    &    \SE{-}     & \SE{0.008}    & \SE{-}           & \SE{-}                   \\ \midrule
\parbox[t]{.2mm}{\multirow{6}{*}{\rotatebox[origin=c]{90}{\Tourism}}}
                                & M         & 0.0910          & \textbf{0.082}    & 0.122            & 0.211           & 0.122             & 0.201          & 0.194         &    0.275      & 0.193         & 0.155            & 0.085                    \\
                                &           & \SE{-}          & \SE{-}            & \SE{-}           & \SE{0.007}      & \SE{0.009}        &  \SE{0.005}    & \SE{0.010}    &    \SE{-}     & \SE{0.004}    & \SE{-}           & \SE{-}                   \\
                                & Q         & 0.099           & \textbf{0.075}    & 0.116            & 0.140           & 0.116             & 0.141          & 0.141         &    0.251      & 0.162         & 0.148            &  0.070                   \\
                                &           & \SE{-}          & \SE{-}            & \SE{-}           & \SE{0.007}      & \SE{0.012}        &  \SE{0.006}    & \SE{0.013}    &    \SE{-}     & \SE{0.0034}   & \SE{-}           & \SE{-}                   \\
                                & Y         & 0.128           & 0.1450            & \textbf{0.116}   & 0.116           & 0.157             &  0.119         & 0.156         &    0.275      & 0.141         & 0.103            & 0.167                    \\
                                &           & \SE{-}          & \SE{-}            & \SE{-}           & \SE{0.011}      & \SE{0.002}        &  \SE{0.011}    & \SE{0.030}    &    \SE{-}     & \SE{0.000}    & \SE{-}           & \SE{-}                   \\ \bottomrule
\end{tabular}}
\end{table}

%% file: sections/04_findings_discussion.tex
\section{Discussion and Conclusion} \label{04_findings_discussion}

Our study covers 15 large-scale datasets, representing a substantial portion of the GIFT-eval collection~\citep{aksu2024giftevalbenchmarkgeneraltime}. In contrast to the recent  GIFT-eval trend of testing methods on artificially extended horizons of the M-series datasets, we deliberately preserve horizons that are consistent with the original Makridakis competitions. The M-series horizons horizons were carefully chosen to reflect the planning needs of practitioners across different domains, and inflating them $10\times$ or $15\times$ beyond their intended range transforms the evaluation into a purely academic exercise, with limited relevance for real-world forecasting applications.

We have conducted a comprehensive evaluation of CFTL. 
Overall, our results serve as a surprising reality check for current claims regarding FFMs.
However, they also point to promising directions for improvement. 
As Appendix~\ref{appendix:synthetic_dataset} shows, augmenting the pretraining datasets with synthetic time series improves \NBEATS's sCRPS performance by 7\%. Similar gains are observed for \MQCNN, \PatchTST, and \Chronos. 
Synthetic data generation is a line of research~\citep{liu2025empoweringtimeseriesanalysis} that will likely be able to bridge the gap between statistical models and FFMs in their CFTL zero-shot regime.

%% file: appendix/a_literature_review.tex
\section{Forecasting Transfer Learning}
\label{appendix:literature_review}

In this section, we summarize the large body of related work on transfer learning for time series forecasting.

\subsection{Single-Frequency Transfer Learning}

Recent advancements in neural forecasting have addressed earlier concerns around computational cost and predictive accuracy, enabling models to consistently outperform traditional statistical approaches~\citep{makriadakis2018concerns}. 
A key driver of this progress is the adoption of cross-learning strategies~\citep{spiliotis2021cross_learning}, where global models are trained on large collections of related time series to extract shared patterns. The cross-learning paradigm underpinned the success of top-performing models in competitions like \Mfour and \Mfive~\citep{smyl2019esrnn, oreshkin2020nbeats}, as well as industry models such as \DeepAR, \MQCNN, \TFT and \SPADE~\citep{salinas2020deepAR, wen2017mqrcnn, lim2021temporal_fusion_transformer,wolff2024spade}.

Transfer learning offers two key practical advantages. 
First, it enables accurate forecasting in scenarios with limited data. 
Second, it streamlines forecasting workflows by reducing the need for extensive model design and hyperparameter tuning, allowing practitioners to obtain strong performance with minimal customization. 
In this sense, transfer learning extends forecasting research agenda initiated by the automation of the Box-Jenkins methodology, which led to models such as \AutoARIMA~\citep{hyndman2008automatic_arima, hyndman2025forecasting}.

The early approaches to transfer learning in time series forecasting focused \EDIT{on one global model per frequency}, 
where success was measured by the model’s ability to outperform traditional statistical baselines---such as \ARIMA, \ETS, and \THeta---in zero-shot settings~\citep{hyndman2008automatic_arima, holt1957exponential_smoothing, fiorucci2016theta, hyndman2025forecasting}. 
In deep learning forecasting literature, this line of research was pioneered by the introduction of meta-learning approach and zero-shot experiments with \NBEATS~\citep{oreshkin2021ts_metalearning}, which laid the groundwork for transfer learning in forecasting. Since then, a series of pre-trained models have emerged, including \TimeGPT~\citep{garza2023timegpt}, \TimesFM~\citep{google2024timesfm}, \LagLlama~\citep{montreal2024lagllama}, and \Chronos~\citep{aws2024chronos}.

\subsection{Cross-Frequency Transfer Learning}

The first attempt to \EDIT{relax the same-frequency constraint in transfer learning} was conducted by Van Ness et al.~\citep{vanness2023crossfrequencytimeseriesmetaforecasting}, 
testing the generalization capabilities of neural forecasting models when the source and target datasets differ in frequency. 
However, their primary results only compared their proposed meta-learning approach, Cross-Frequency Adapter (\CFA), and other neural forecasting models such as \LSTM and \NBEATS. 
Their evaluation left unanswered the critical question of whether CFTL outperforms traditional statistical baselines.

Woo et al.~\citep{salesforce2023moirai}, introduced \MOIRAI, a Universal Time Series Forecasting model capable of cross-frequency transferability. 
By pretraining on their LOTSA dataset, \MOIRAI claims that CFTL improved upon fully trained neural forecasting models and statistical baselines. 
While the paper's primary focus is on long-horizon forecasting tasks, they report aggregated results from the Monash Time Series Forecasting Benchmark~\citep{godahewa2021monash_repository}, using the normalized Mean Absolute Error (nMAE) as the evaluation metric. 
In these evaluations, \MOIRAI\ claimed to achieve relative improvements over \THeta, \ARIMA, and \ETS, by an average of 38\%, 36\%, and 35\%, and 15\% upon fully trained \NBEATS. A revision of \MOIRAI's Table 20 on disaggregated evaluation on the Monash repository revealed suspiciously volatile measurements where they improve performance by 94\% upon \ETS on \Mfour-hourly, while degrade performance by 77.24\% on \Tourism-Quarterly.
This raises questions on the execution of their statistical baselines.

In a parallel line of work, Fatir et al. introduced \Chronos, a model also designed to perform CFTL. 
In their experiments, they evaluate \Chronos's zero-shot accuracy across 27 datasets, including the M-forecasting competitions, Tourism and Dominick datasets, as well as long-horizon datasets~\citep{zhou2021informer}. 
With sCRPS measures, \Chronos asserts improved average performance upon Theta, ARIMA, and ETS by 47\%, 35\%, and 47\%. 
A potential issue with the statement of their performance gains lies in the uniformly averaged performance calculation across datasets; such a reporting is convenient and common, but it disproportionately skews the measurements towards the smaller datasets like long-horizon~\citep{zhou2021informer}.

%% file: appendix/b_dataset_details.tex
\section{Dataset Details}
\label{appendix:dataset_details}

In this section, we provide a summary of the data we used in our evaluation.

\subsection{Pre-Training Datasets}

\begin{figure}[t] 
    \centering
    \subfigure[Sample from Pretraining Dataset]{
    \label{fig:internal_dataset}
    \includegraphics[width=0.46\linewidth]{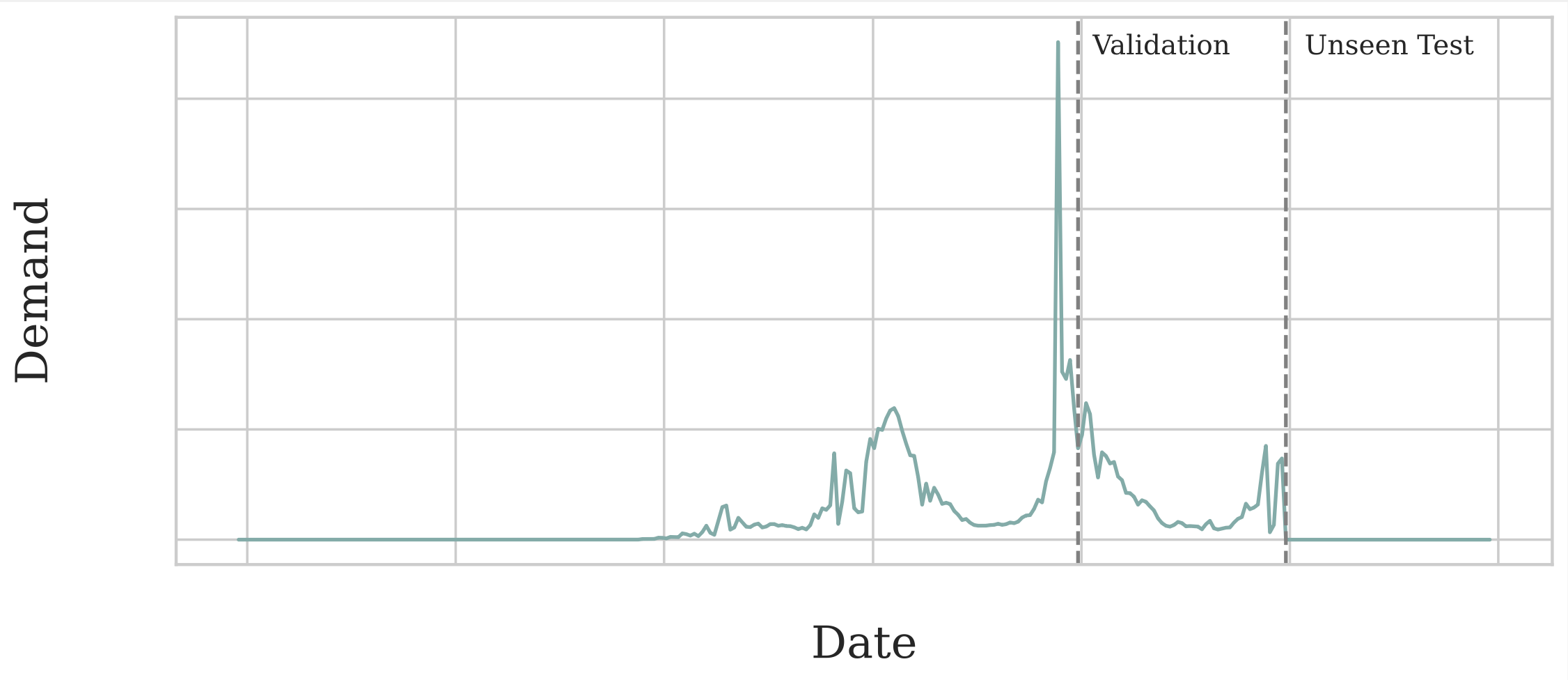}
    }
    \subfigure[Synthetic Dataset]{
    \label{fig:synthetic_dataset}
    \includegraphics[width=0.39\linewidth]{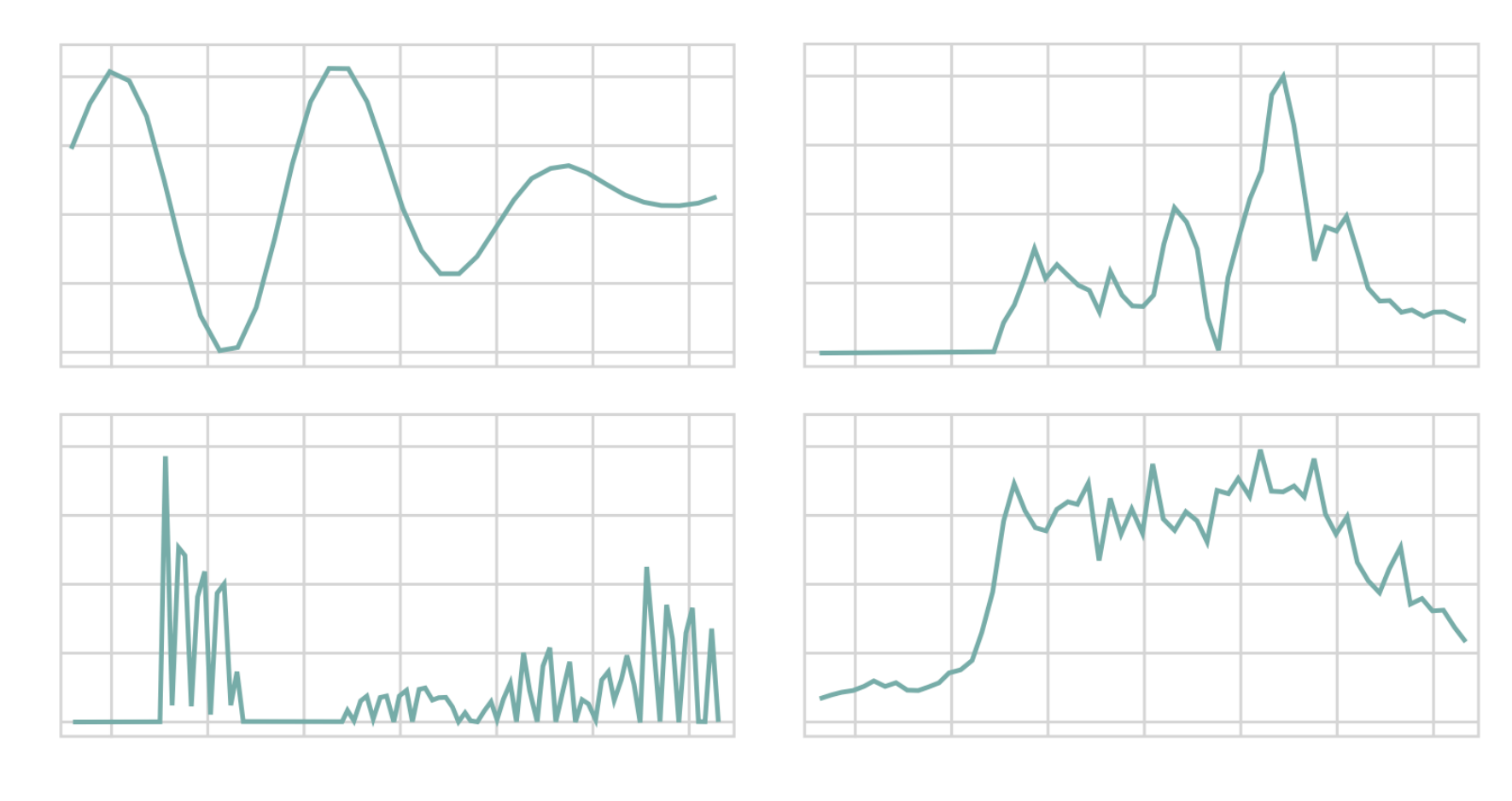}
    }
    \caption{For our CFTL task, we use two datasets: (a) a set of real-world datasets composed of large-scale online retail demand; and (b) a set of synthetic dataset composed of Gaussian processes, Fourier harmonic signals, and polynomial trends. 
    }
    \label{fig:training_methodology}
\end{figure}

Here, we describe the datasets we used in our pre-training. 
See Table~\ref{table:amazon_datasets} 
and Figure~\ref{fig:training_methodology}
for a summary.

\paragraph{Real-world data.}
The primary source of data for our pre-training consisted of several real-world datasets, which we summarize here.

\input{tables/amazon_datasets.tex}

\GO comes from a chain of convenience stores operating in multiple countries. The dataset contains demand data for various consumer products including food items. 

\Fresh represents daily demand from a grocery delivery service operating in multiple regions globally. The service offers various food and household products to subscribers. 

\Mendel originates from a hybrid retail format that combines multiple fulfillment methods. It includes daily demand data from stores in North America, supporting both in-store shopping and delivery options. 

\FBA contains weekly demand data from a third-party fulfillment service operating across six developed countries. The service handles all aspects of product storage and delivery for external sellers. 

\GlobalRetail comprises national-level demand data from a major retail platform, including information from multiple countries around the world. \ZIPtwo is a more granular version of \GlobalRetail's data for one country, broken down by postal code prefixes. It shows more irregular demand patterns than the national-level data. 


\clearpage
\paragraph{Synthetic Datasets.}
We also used carefully-constructed synthetic data for pre-training.

Dataset G was artificially generated to supplement the training data, incorporating various time series patterns. 
These include basic constants, sinusoidal and cosinusoidal seasonalities, linear trends, polynomial trends, frequency drift curves, gaussian waves, exponential trend, and logistic growth curves across different time granularities as seen in Figure~\ref{fig:synthetic_dataset}. The patterns were combined and modified with random noise to create realistic variations. The basic component signal equations are provided below: 

\begin{gather*}
\label{evaluation_metrics}
\begin{alignedat}{3}
        \mathbf{y}_{t}&=k                                                 & \qquad\qquad
        \mathbf{y}_{t}&=\mathrm{sin}(\pi * a * t + b)                     & \qquad\qquad
        \mathbf{y}_{t}&=\mathrm{cos}(\pi * a * t + b)                     \\
        &&\\
        \mathbf{y}_{t}&=a * t + b                                         & \qquad\qquad
        \mathbf{y}_{t}&=\mathrm{sin}(\frac{\pi * a}{t} + b)               & \qquad\qquad
        \mathbf{y}_{t}&=a * \mathrm{exp}\left(-\frac{(t-b)}{c}^{2}\right) \\
        &&\\
        \mathbf{y}_{t}&=a * \mathrm{exp}\left(b * t\right)                & \qquad\qquad
        \mathbf{y}_{t}&=a * t^{2} + b * t + c                             & \qquad\qquad
        \mathbf{y}_{t}&=\frac{a}{1 + \mathrm{exp}\left(-b * (t-c)\right)} \\
\end{alignedat}
\end{gather*}

\clearpage
\subsection{Evaluation Datasets}

Here, we describe the datasets we used in our evaluation.
See Table~\ref{table:datasets} for a summary.

\input{tables/evaluation_datasets.tex}

\paragraph{M1 Dataset Details.}
The early \Mone competition~\citep{makridakis1982m1_competition}, organized by Makridakis et al., focused on 1,001 time series drawn from demography, industry, and economics, with lengths ranging from 9 to 132 observations and varying in frequency (monthly, quarterly, and yearly). A key empirical finding of this competition was that simple forecasting methods, such as ETS~\citep{holt1957exponential_smoothing}, often outperformed more complex approaches. These results had a lasting impact on the field, initiating a research legacy that emphasized accurate forecasting, model automation, and caution against overfitting. The competition also marked a conceptual shift, helping to distinguish time-series forecasting from traditional time series analysis.

\paragraph{M3 Dataset Details.}
The \Mthree competition~\citep{makridakis2000m3_competition}, held two decades after the \Mone competition, featured a dataset of 3,003 time series spanning business, demography, finance and economics. These series ranged from 14 to 126 observations and included monthly, quarterly, and yearly frequencies. All series had positive values, with only a small proportion displaying erratic behavior and none exhibiting intermittency~\citep{syntetos2005syntetos_categorization}. The \Mthree competition reinforced the trend of simple forecasting methods outperforming more complex alternatives, with the \THeta method~\citep{hyndman2003unmasking_theta} emerging as the best performing approach.

\paragraph{M4 Dataset Details.}
The \Mfour competition marked a substantial increase in both the size and diversity of the M competition datasets, comprising 100,000 time series across six frequencies: hourly, daily, weekly, monthly, quarterly, and annual. These series covered a wide range of domains, including demography, finance, industry, and both micro- and macroeconomic indicators. The competition also introduced the evaluation of prediction intervals in addition to point forecasts, broadening the assessment criteria. \Mfour's proportion of non-smooth or erratic time series increased to 18 percent~\citep{syntetos2005syntetos_categorization}. For the first time, a neural forecasting model - ESRNN\citep{smyl2019esrnn} - outperformed traditional methods. The competition also helped popularize cross-learning~\citep{spiliotis2021cross_learning} in global models.

\paragraph{Tourism Dataset Details.}
The Tourism dataset~\citep{athanosopoulus2011tourism_competition} was designed to evaluate forecasting methods applied to tourism demand data across multiple temporal frequencies. It comprises 1,311 time series at monthly, quarterly, and yearly frequencies. This competition introduced the Mean Absolute Scaled Error (MASE) as an alternative metric to evaluate scaled point forecasts, alongside the evaluation of forecast intervals. Notably, 36\% of the series were classified as erratic or intermittent. Due to this high proportion of irregular data, the Naïve1 method proved particularly difficult to outperform at the yearly frequency.

%% file: tables/amazon_datasets.tex
\begin{table}[!ht]
\caption{Summary of forecasting datasets used 
for pre-training. 
} 
\label{table:amazon_datasets}
\centering
\footnotesize
    \begin{tabular}{llllcccc}
    \toprule
                                            & Frequency & Horizon  & Series      & Min Length & Max-Length \\ \midrule
    \multirow{2}{*}{\GO}                    & Daily     & 24       & 65K         & 1          & 1857       \\
                                            & Daily     & 24       & 28K         & 1          & 1857       \\ \midrule
    \multirow{1}{*}{\FBA}                   & Weekly    & 24       & 4MM         & 1          & 262        \\ \midrule
    \multirow{1}{*}{\Mendel}                & Daily     & 24       & 900K        & 1          & 1857       \\ \midrule
    \multirow{3}{*}{\Fresh}                 & Daily     & 24       & 350K        & 1          & 1857       \\ 
                                            & Daily     & 24       & 320K        & 1          & 1857       \\
                                            & Daily     & 24       & 1.5MM       & 1          & 1857       \\ \midrule
    \multirow{3}{*}{\GlobalRetail}          & Daily     & 24       & 290MM       & 1          & 1834       \\
                                            & Weekly    & 24       & 290MM       & 1          & 262        \\
                                            & Monthly   & 24       & 290MM       & 1          & 58         \\ \midrule
    \multirow{1}{*}{\ZIPtwo}                & Weekly    & 24       & 700MM       & 1          & 314        \\                                                   
    \bottomrule
    \end{tabular}
\end{table}

%% file: tables/evaluation_datasets.tex
\begin{table}[!ht]
\caption{Summary of forecasting datasets 
we used in our evaluation. 
} 
\label{table:datasets}
\centering
\footnotesize
\begin{tabular}{llcccccc}
\toprule
                         & Frequency & Seasonality & Horizon                     & Series                     & Min Length                     & Max Length                     & \% Erratic                     \\ \midrule
\multirow{3}{*}{M1}      & Monthly   & 12          & 18                          & 617                        & 48                             & 150                            & 0                              \\
                         & Quarterly & 4           & 8                           & 203                        & 18                             & 114                            & 0                              \\
                         & Yearly    & 1           & 6                           & 181                        & 15                             & 58                             & 0                              \\ \midrule
\multirow{4}{*}{M3}      & Other     & 4           & 8                           & 174                        & 71                             & 104                            & 0                              \\
                         & Monthly   & 12          & 18                          & 1428                       & 66                             & 144                            & 2                              \\
                         & Quarterly & 4           & 8                           & 756                        & 24                             & 72                             & 1                              \\
                         & Yearly    & 1           & 6                           & 645                        & 20                             & 47                             & 10                             \\ \midrule
\multirow{6}{*}{M4}      & Hourly    & 24          & 48                          & 414                        & 748                            & 1008                           & 17                             \\ 
                         & Daily     & 1           & 14                          & 4,227                      & 107                            & 9933                           & 2                              \\
                         & Weekly    & 1           & 13                          & 359                        & 93                             & 2610                           & 16                             \\
                         & Monthly   & 12          & 18                          & 48,000                     & 60                             & 2812                           & 6                              \\
                         & Quarterly & 4           & 8                           & 24,000                     & 24                             & 874                            & 11                             \\
                         & Yearly    & 1           & 6                           & 23,000                     & 19                             & 841                            & 18                             \\ \midrule
\multirow{3}{*}{Tourism} & Monthly   & 12          & 24                          & 366                        & 91                             & 333                            & 51                             \\
                         & Quarterly & 4           & 8                           & 427                        & 30                             & 130                            & 39                             \\
                         & Yearly    & 1           & 4                           & 518                        & 11                             & 47                             & 23                             \\ \bottomrule
\end{tabular}
\end{table}

%% file: appendix/b_point_forecast.tex
\section{Point Forecast Results}
\label{appendix:point_forecast}

In this section, we complement our evaluation of probabilistic forecasts (from the main text) with a set of point forecast results.
We consider the \emph{Mean Absolute Scaled Error} (MASE, \citep{hyndman2006another_look_measures}), that considers the ratio between mean absolute error of forecasts over mean absolute error of the \Naive forecast $\tilde{y}_{i,t,h}$ (i.e., a point forecast using the last observation on the previous season), as described by
\begin{equation}
    \mathrm{MASE}\left(\mathbf{y},\; \hat{\mathbf{y}},\; \tilde{\mathbf{y}} \right)  = 
    \frac{\sum_{i,t,h}|y_{i,t,h}-\hat{y}_{i,t,h}|}
    {\sum_{i,t,h}|y_{i,t,h}-\tilde{y}_{i,t,h}|} .
\end{equation}

\input{tables/main_mase.tex}

Table \ref{table:main_mase} reports mean point forecast performances of our statistical baselines and neural forecast models using the MASE across the five best checkpoints during the training process.
The lowest value in every dataset-frequency cell again belongs to a statistical baseline method.
On \Mthree-Other, \SCUM reaches a MASE of 0.515 against \Chronos-P's 0.637; on \Mfour-Daily, the exponential-smoothing family (\ARIMA, \THeta, \ETS) MASE ranges from 0.963-0.977 while the best neural forecasts range from 3.000-3.330.

While gaps are smaller for lower frequency data, classical models still lead: \CES reports a MASE of 0.636 on \Mone-Quarterly compared to the zero-shot \Chronos-P network's MASE of 0.766, and \THeta has a MASE of 0.625 on \Tourism-Quarterly versus the 1.332  of \MQCNN;
neural forecasters never obtain the minimum MASE in any dataset or frequency, and exceed 1.0 in most rows.
On the other hand, \SCUM, (Ensemble of \CES, \ETS, \THeta, and \ARIMA) stay below the 1.0 threshold in all but a few yearly series.

These results mirror our results for probabilistic scores; and they confirm that, also for point forecasts, traditional statistical like \ARIMA, and \SCUM methods remain the most accurate choice on the four benchmark suites.


%% file: tables/main_mase.tex
\begin{table}[!t]
\caption{Empirical evaluation of point forecasts. Mean \emph{absolute scaled error} (MASE) averaged over 5 runs. The best united CFTL framework result is highlighted (lower measurements are preferred).
The methods without standard deviation have deterministic solutions.\\
\tiny{
\tik \ChronosBS\tik stands for a pretrained \texttt{Chronos-Bolt-Small}. Zero-shot predictions correspond to the original Hugging face model published by Fatir et al~\citep{aws2024chronos}.
\tik \Chronos was, trained in our unified CFTL framework, without being full-shot we are able to replicate or improve \ChronosBS accuracy in various datasets.\\
\tik\tik Neither \TimesFM nor \ChronosBS are zero-shot forecasting models as they are trained on the \Mfour dataset~\citep{aws2024chronos, google2024timesfm}.
}
} \label{table:main_mase}
\centering
\scriptsize
\resizebox{\textwidth}{!}{\begin{tabular}{ll|cc|ccccc|cccc}
\toprule
                              &       & \multicolumn{2}{|c}{StatsForecast}               &  \multicolumn{5}{|c}{NF (unified CFTL framework)}                                       & \multicolumn{4}{|c}{NF (external train)}                                \\ \midrule
                              & Freq  & \ARIMA                & \SCUM                    & Best             &  \NBEATS         & \MQCNN          & \PatchTST     & \Chronos\tik    & \MOIRAI-S        & \TabPFN         & \ChronosBS\tik  & \TimesFM\tik\tik \\ \midrule
\multirow{6}{*}{\Mone}        & M     & 0.759                 & 0.765                    & \textbf{0.715}   &  0.896           & 0.745           & 0.715         & 1.048           &   0.659          & 0.838           & 0.834           & 0.655            \\
                              &       & \SE{-}                & \SE{-}                   & \SE{-}           &  \SE{0.039}      & \SE{0.002}      & \SE{0.007}    & \SE{0.018}      &   \SE{-}         & \SE{-}          & \SE{-}          & \SE{-}           \\
                              & Q     & 0.889                 & 0.801                    & \textbf{0.699}   &  1.026           & 0.791           & 0.707         & 1.078           &   0.778          & 0.972           & 0.818           &  1.039           \\
                              &       & \SE{-}                & \SE{-}                   & \SE{-}           &  \SE{0.176}      & \SE{0.007}      & \SE{0.028}    & \SE{0.125}      &   \SE{-}         & \SE{-}          & \SE{-}          & \SE{-}           \\
                              & Y     & 0.718                 & 0.686                    & \textbf{0.632}   &  0.672           & 0.977           & 0.629         & 0.993           &   1.289          & 0.830           & 0.723           &  0.803           \\ 
                              &       & \SE{-}                & \SE{-}                   & \SE{-}           &  \SE{0.092}      & \SE{0.012}      & \SE{0.009}    & \SE{0.072}      &   \SE{-}         & \SE{-}          & \SE{-}          & \SE{-}           \\ \midrule
\multirow{8}{*}{\Mthree}      & O     & 0.738                 & \textbf{0.693}           & 0.784            &  1.040           & 0.968           & 0.822         & 0.784           &    0.725         & 0.866           & 0.729           &  0.853           \\
                              &       & \SE{-}                & \SE{-}                   & \SE{-}           &  \SE{0.592}      & \SE{0.021}      & \SE{0.165}    & \SE{0.143}      &    \SE{-}        & \SE{-}          & \SE{-}          & \SE{-}           \\
                              & M     & 0.775                 & \textbf{0.721}           & 0.795            &  0.861           & 0.888           & 0.795         & 0.860           &    0.936         & 0.838           & 0.880           &  0.709           \\
                              &       & \SE{-}                & \SE{-}                   & \SE{-}           &  \SE{0.150}      & \SE{0.002}      & \SE{0.002}    & \SE{0.006}      &    \SE{-}        & \SE{-}          & \SE{-}          & \SE{-}           \\
                              & Q     & 0.905                 & \textbf{0.821}           & 0.856            &  0.959           & 0.937           & 0.852         & 1.394           &    1.008         & 0.941           & 0.879           &  0.882           \\
                              &       & \SE{-}                & \SE{-}                   & \SE{-}           &  \SE{0.252}      & \SE{0.005}      & \SE{0.032}    & \SE{0.083}      &    \SE{-}        & \SE{-}          & \SE{-}          & \SE{-}           \\
                              & Y     & 1.104                 & 0.998                    & \textbf{0.736}   &  0.887           & 1.212           & 0.743         & 1.141           &   1.045          & 0.957           & 0.841           &  1.023           \\
                              &       & \SE{-}                & \SE{-}                   & \SE{-}           &  \SE{0.106}      & \SE{0.019}      & \SE{0.038}    & \SE{0.070}      &    \SE{-}        & \SE{-}          & \SE{-}          & \SE{-}           \\ \midrule
\multirow{10}{*}{\Mfour}      & D     & 0.977                 & \textbf{0.962}           & 1.041            &  3.007           & 1.041           & 0.847         & 0.974           &    1.323         & 1.055           & 1.087           &  0.965           \\
                              &       & \SE{-}                & \SE{-}                   & \SE{-}           &  \SE{0.196}      & \SE{0.192}      & \SE{0.008}    & \SE{0.030}      &    \SE{-}        & \SE{-}          & \SE{-}          & \SE{-}           \\
                              & W     & 0.886                 & 0.931                    & \textbf{0.861}   &  1.281           & 0.861           & 0.890         & 1.128           &    1.378         & 0.903           & 1.004           &  0.814           \\
                              &       & \SE{-}                & \SE{-}                   & \SE{-}           &  \SE{0.061}      & \SE{0.063}      & \SE{0.009}    & \SE{0.014}      &    \SE{-}        & \SE{-}          & \SE{-}          & \SE{-}           \\
                              & M     & 0.839                 & \textbf{0.811}           & 0.864            &  0.898           & 0.911           & 0.800         & 0.864           &    1.102         & 0.895           & 0.948           & 0.605            \\
                              &       & \SE{-}                & \SE{-}                   & \SE{-}           &  \SE{0.016}      & \SE{0.005}      & \SE{0.007}    & \SE{0.005}      &    \SE{-}        & \SE{-}          & \SE{-}          & \SE{-}           \\
                              & Q     & 0.874                 & \textbf{0.838}           & 0.936            &  0.936           & 0.970           & 0.795         & 1.082           &    1.234         & 0.953           & 0.887           & 0.695            \\
                              &       & \SE{-}                & \SE{-}                   & \SE{-}           &  \SE{0.019}      & \SE{0.072}      & \SE{0.015}    & \SE{0.114}      &    \SE{-}        & \SE{-}          & \SE{-}          & \SE{-}           \\
                              & Y     & 0.921                 & \textbf{0.814}           & 0.944            &  0.944           & 1.043           & 0.700         & 0.988           &    1.464         & 0.924           & 0.789           & 0.667            \\
                              &       & \SE{-}                & \SE{-}                   & \SE{-}           &  \SE{0.093}      & \SE{0.118}      & \SE{0.046}    & \SE{0.128}      &    \SE{-}        & \SE{-}          & \SE{-}          & \SE{-}           \\ \midrule
                              & M     & 0.368                 & \textbf{0.333}           & 0.509            &  0.881           & 0.509           & 0.865         & 0.842           &    1.148         & 0.860           & 0.636           & 0.357            \\
                              &       & \SE{-}                & \SE{-}                   & \SE{-}           &  \SE{0.018}      & \SE{0.007}      & \SE{0.019}    & \SE{0.044}      &    \SE{-}        & \SE{-}          & \SE{-}          & \SE{-}           \\
                              & Q     & 0.727                 & \textbf{0.539}           & 0.843            &  1.026           & 0.843           & 0.912         & 1.105           &    1.840         & 1.142           & 1.028           & 0.539            \\
                              &       & \SE{-}                & \SE{-}                   & \SE{-}           &  \SE{0.059}      & \SE{0.007}      & \SE{0.032}    & \SE{0.105}      &    \SE{-}        & \SE{-}          & \SE{-}          & \SE{-}           \\
                              & Y     & 0.744                 & 0.791                    & \textbf{0.577}   &  0.672           & 0.880           & 0.588         & 0.773           &     1.558        & 0.842           &  0.562          & 0.924            \\
                              &       & \SE{-}                & \SE{-}                   & \SE{-}           &  \SE{0.092}      & \SE{0.012}      & \SE{0.052}    & \SE{0.159}      &    \SE{-}        & \SE{-}          & \SE{-}          & \SE{-}           \\ \bottomrule
\end{tabular}}
\end{table}

%% file: appendix/c_training_methodology_hypars.tex
\section{Training Methodology and Hyperparameters}
\label{appendix:training_methodology}


\input{tables/hyperparameters.tex}

\vspace{-.5 cm}
In this section, we provide details on the training methodology, outlined in Section~\ref{03_experiments}. 
The optimization of all models is based on the definition of training, validation, and test datasets, depicted in Figure~\ref{fig:training_methodology}. 
For all our pre-training datasets, we keep the 24 observations immediately following the training data as validation. 
Given the scale of our evaluation, we focused our hyperparameter optimization solely on the selection of training steps and learning rate, and we rely principally on the default hyperparameters implementation for each baseline. 
See 
Tables~\ref{table:hyperparameters_nbeats},
\ref{table:hyperparameters_mqcnn},
 \ref{table:hyperparameters_patchtst}, \ref{table:hyperparameters_chronos}.
 Hyperparameters not specified in these tables are set to the defaults of the original implementations in the NeuralForecast library~\citep{olivares2022neuralforecast}, or the Chronos repository~\citep{aws2024chronos}.

We conducted all neural network experiments using a single AWS p4d.24xlarge	with 1152 GiB of RAM and 96 vCPUs. Training times mostly depend on the architecture, however we restrict the SGD training steps to 100K per architectures. 

%% file: tables/hyperparameters.tex
\begin{table}[!htb]
\label{table:hyperparameters}
    \begin{minipage}{.5\linewidth}
      \caption{\NBEATS} \label{table:hyperparameters_nbeats}
      \centering
      \scriptsize
        \begin{tabular}{lc}
        \toprule
        \textsc{Hyperparameter}                                         & \textsc{Values}           \\ \midrule
        Single GPU SGD Batch Size\textsuperscript{*}.                   & 32 (32*8)                 \\
        Initial learning rate.                                          & 0.001                     \\
        Maximum Training steps $S_{max}$.                               & 60,000                    \\ 
        Learning rate decay.                                            & 0.1                       \\
        Learning rate steps.                                            & 40,000; 50,000            \\ \midrule
        Input size.                                                     & 48                        \\
        Main Activation Function.                                       & ReLU                      \\
        Number of Stacks                                                & 4                         \\
        Number of Blocks within Stacks.                                 & 3                         \\
        MLP layers within Blocks.                                       & 2                         \\
        Coefficients hidden size.                                       & 512                       \\
        Degree of Trend Polynomials (interpretable).                    & N/A                       \\
        Number of Fourier Basis (interpretable).                        & N/A                       \\
        \bottomrule
        \end{tabular}
    \end{minipage}%
\hspace{1.cm}
    \begin{minipage}{.5\linewidth}
      \caption{\MQCNN} \label{table:hyperparameters_mqcnn}
      \centering
      \scriptsize
        \begin{tabular}{lc}
        \toprule
        \textsc{Hyperparameter}                                         & \textsc{Values}                \\ \midrule
        Single GPU SGD Batch Size\textsuperscript{*}.                   & 32 (32*8)                      \\
        Initial learning rate.                                          & 0.001                          \\
        Maximum Training steps $S_{max}$.                               & 400,000                        \\ 
        Learning rate decay.                                            & 0.1                            \\
        Learning rate steps.                                            & 400,000 / 2                    \\ \midrule
        Main Activation Function                                        & {ReLU}                         \\
        Temporal Convolution Kernel Size                                & {2}                            \\
        Temporal Convolution Dilations.                                 & {$[1,2,4,8,16,32]$}            \\
        Historic Encoder Dimension.                                     & {30}                           \\
        Future Encoder Dimension ($hf_{1}$).                            & {50}                           \\
        Static Encoder D.Multip. $(\alpha\times|\sqrt{x}^{(s)}|)$       & {30}                           \\
        H-Agnostic Decoder Dimension.                                   & {100}                          \\
        H-Specific Decoder Dimension.                                   & {20}                           \\
        \bottomrule
        \end{tabular}
    \end{minipage}%
\end{table}

\begin{table}[!htb]
\label{table:hyperparameters2}
\label{table:hyperparameters3}
    \begin{minipage}{.5\linewidth}
      \caption{\PatchTST} \label{table:hyperparameters_patchtst}
      \centering
      \scriptsize
        \begin{tabular}{lc}
        \toprule
        \textsc{Hyperparameter}                                         & \textsc{Values}           \\ \midrule
        Single GPU SGD Batch Size\textsuperscript{*}.                   & 32 (32*8)                 \\
        Initial learning rate.                                          & 0.001                     \\
        Maximum Training steps $S_{max}$.                               & 100,000                   \\ 
        Learning rate decay.                                            & 0.1                       \\
        Learning rate steps.                                            & 100,000 / 5               \\ \midrule
        Input Size.                                                     & 128                       \\
        Main Activation Function                                        & ReLU                      \\
        Patching Length.                                                & 16                        \\
        Patching Stride.                                                & 8                         \\
        Number of Attention Heads.                                      & 16                        \\
        Encoder Hidden Size.                                            & 128                       \\        
        Decoder Hidden Size.                                            & 256                       \\
        Apply Revin.                                                    & True                      \\
        Residualized Attention.                                         & True                      \\
        \bottomrule
        \end{tabular}
    \end{minipage}%
\hspace{1.cm}
\vspace{1.cm}
    \begin{minipage}{.5\linewidth}
      \caption{\ChronosBolt} \label{table:hyperparameters_chronos}
      \centering
      \scriptsize
        \begin{tabular}{lc}
        \toprule
        \textsc{Hyperparameter}                                         & \textsc{Values}           \\ \midrule
        Single GPU SGD Batch Size\textsuperscript{*}.                   & 4 (96)                    \\
        Initial learning rate.                                          & 0.0005                    \\
        Maximum Training steps $S_{max}$.                               & 50,000                    \\ 
        Learning rate decay.                                            & 0.1                       \\
        Learning rate steps.                                            & 50,000 / 5                \\ \midrule
        Input Size.                                                     & 2048                      \\
        Main Activation Function                                        & {ReLU}                    \\
        Encoder/Decoder Hidden Size.                                    & 256                       \\
        Encoder Type.                                                   & \texttt{T5Stack}          \\
        Decoder Type.                                                   & \texttt{T5Stack}          \\
        Patch Size                                                      & 16                        \\
        Patch Stride                                                    & 16                        \\
        Encoder Number of Layers.                                       & 4                         \\
        Decoder Number of Layers.                                       & 4                         \\
        Number of Attention Heads.                                      & 4                         \\
        Attention Dropout Rate.                                         & 0.1                       \\
        \bottomrule
        \end{tabular}
    \end{minipage}%
\end{table}

%% file: appendix/d_note_on_statistical_ensemble.tex
\section{Implementation Details of the Simple Combination of Univariate Models}
\label{appendix:statistical_ensemble}

In this section, we provide details on the implementation of the statistical ensemble used to generate the point and probabilistic forecasts evaluated in Table~\ref{table:main_crps} and Table~\ref{table:main_mase}.

As discussed in Section~\ref{03_experiments}, we employ the Simple Combination of Univariate Models (\SCUM; \citep{petropoulos2020scum}) framework. This ensemble method aggregates forecasts from four classical statistical models. Complex Exponential Smoothing (\CES; \citep{svetunkov2022ces}), Dynamic Optimized Theta (\THeta; \citep{fiorucci2016theta}), Automatic Autoregressive Integrated Moving Average (\ARIMA; \citep{hyndman2008automatic_arima}), and Exponential Smoothing (\ETS; \citep{holt1957exponential_smoothing}). 
For all the models, we use the implementations of the StatsForecast library~\citep{garza2022statsforecast, hyndman2025forecasting}.

Each model is independently fitted to the time series, producing Gaussian-distributed forecasts. 
Assuming Normality and independence among model forecast distributions, we construct the ensemble by aggregating the means and variances of the individual forecasts. 
Let \CES, \THeta, \ARIMA, and \ETS denote the constituent models, the ensemble forecast is computed as:

\begin{equation}
\begin{aligned}
\hat{\mu} = \frac{1}{4}\left(\hat{\mu}_{\mathrm{CES}} + \hat{\mu}_{\mathrm{Theta}} + \hat{\mu}_{\mathrm{ARIMA}} + \hat{\mu}_{\mathrm{ETS}}\right)
\end{aligned}
\end{equation}

\begin{equation}
\begin{aligned}
\hat{\sigma}^{2} = \frac{1}{4}\left(\hat{\sigma}^{2}_{\mathrm{CES}} + \hat{\sigma}^{2}_{\mathrm{Theta}} + \hat{\sigma}^{2}_{\mathrm{ARIMA}} + \hat{\sigma}^{2}_{\mathrm{ETS}}\right)
\end{aligned}
\end{equation}

We generate the final quantile predictions using the percent point function:
\begin{equation}
\begin{aligned}
\hat{y}^{(q)} &= \hat{\mu} + \hat{\sigma} z^{(q)} \\
\qquad \mathrm{ with }\qquad  
z^{(q)} &= \mathrm{inf}\{y \in \mathbb{R} \;:\; q\leq \Phi(y) \}
\end{aligned}
\end{equation}

To run the statistical baselines we used a single AWS c5.18xlarge instance with 72 vCPUs and 137 GiB of RAM. To ensure the reproducibility of our experimental results, we provide the implementation of the statistical baselines at the following link: \url{https://anonymous.4open.science/r/neurips_baselines-4BC5}.

%% file: appendix/e_ablation_studies.tex
\section{Ablation Study on Synthetic Data in our Pre-training Datasets }
\label{appendix:synthetic_dataset}

\begin{figure}[!ht] 
    \centering
    \label{fig:synthetic_data_ablationm1}
    \includegraphics[width=0.45\linewidth]{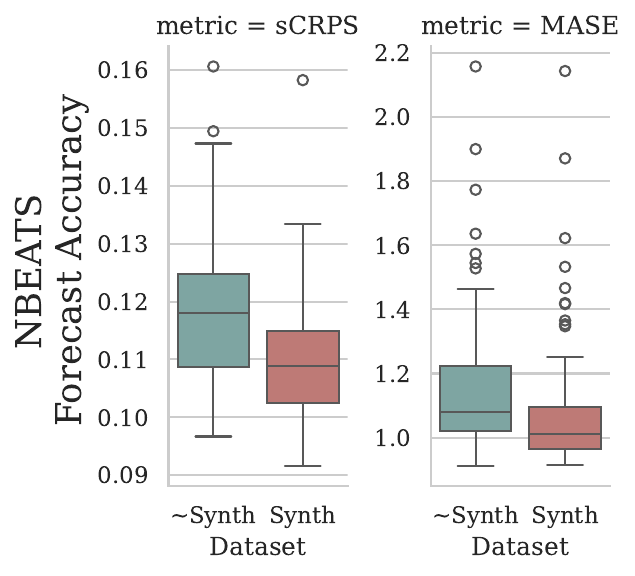}
    \caption{Pre-training datasets ablation, with and without the use of synthetic data.  Shown are metrics with (red) and without (green) synthetic data for pre-training, for the \NBEATS model. 
    }
    \label{fig:ablation_studies}
\end{figure}

Our re-implementation of well-established univariate forecasting algorithms, adapted for the CFTL task, enabled us to isolate a primary driver of accuracy improvements across architectures: dataset quality. 
As shown in Figure~\ref{fig:ablation_studies}, our CFTL-adapted \NBEATS model improved its sCRPS score from 0.116 to 0.108 - a notable 7\% gain - when synthetic data was added to the pre-training set. 
Similar improvements were observed across other architectures.
For this and other models, our results demonstrated that dataset composition, rather than architectural choices, was the primary driver of sCRPS improvements.

Importantly, even in the presence of huge pre-training datasets, of 1.58 billion series, synthetic data are still capable of improving the zero-shot performance of \NBEATS, \MQCNN, \Chronos, and \PatchTST (as shown in Table~\ref{table:main_crps} and Table~\ref{table:main_mase}), reinforcing the central role of training data in model performance even at large scales. 
This suggests that better synthetic data generation methodologies will be important to the future advancements of CFTL and FFMs.